\documentclass[conference]{IEEEtran}
\IEEEoverridecommandlockouts
\usepackage{cite}
\usepackage{amsmath,amssymb,amsfonts}
\usepackage{algorithm2e}
\usepackage{graphicx}
\usepackage{textcomp}
\usepackage{xcolor}
\usepackage{graphicx}
\usepackage{caption}
\usepackage{adjustbox}
\usepackage{amsmath}
\usepackage{lipsum}
\usepackage{amsfonts}
\def\BibTeX{{\rm B\kern-.05em{\sc i\kern-.025em b}\kern-.08em
    T\kern-.1667em\lower.7ex\hbox{E}\kern-.125emX}}

\begin{document}

\title{Agree To Disagree\\}

\author{\IEEEauthorblockN{Abhinav Raghuvanshi}
\IEEEauthorblockA{\textit{Department Of Aerospace Engineering} \\
\textit{Indian Institute of Technology, Bombay}\\
Mumbai, India\\
200040008@iitb.ac.in}

\and
\IEEEauthorblockN{Anirudh Mittal}
\IEEEauthorblockA{\textit{AI Resident} \\
\textit{Meta}\\
London, UK \\
anirudhmittal@meta.com}

\and
\IEEEauthorblockN{Siddhesh Pawar}
\IEEEauthorblockA{\textit{Researcher} \\
\textit{Google AI}\\
Bengaluru, India \\
siddheshmp@google.com}
}

\maketitle

\begin{abstract}
How often do you read the terms and conditions before actually signing up for something, installing a software or entering a website? Most internet users don’t. The reason behind that shouldn't be very hard to find. The Terms and conditions are usually many pages long filled with legal jargon and complex sentences. Through this paper we present a Machine Learning based method that will read the document for you and summarize the important points in simple language that actually matter to you and which you might want to consider before signing up.

\end{abstract}


\section{Introduction}
In the ever growing era of networking and proportionately swift upcoming of new websites everyday has led to creation of newer methods for providing better user experience on a website that might involve usage of user data. A website Privacy policy is a statement or a legal document that discloses some or all of the ways a party (the website) gathers, uses, discloses and manages a customer or client's data. Every person has the right to privacy which means he is the owner of his own data and can choose what they wish to share and with whom. As per the Law every organization that has control over its users data is obligated to maintain standards for complete data privacy and security, but everyday users do not posses a modus operandi to know what kind of data collection does the website do in the background or what does it exactly do with the data. Even if one out of a many users do try to navigate through to the 'Terms of Service' Page, most people do not posses the technical or legal knowledge to understand what does the long and detailed policies mean. 

Ideally every policy should present data in a human understandable way and should clearly mention how they plan to collect,store and use it but all the details seem to be shadowed by the complicated language websites use to unknowingly make user take the easier way out by not bothering what lies between the lines of the text.

To assist the user in making informed and smarter choices by not only evaluating the text in the policies but also showing meaningful summaries of the policies, we aim to provide a smoother and safer web surfing experience for people. We perform a complete comprehensive analysis of each part of the policy mentioned on the website and the algorithms could be modified easily for 'Cookie Policies', 'Terms of Service' and/or 'Privacy Policies'. We not only analyse the text, we have also implemented a scoring algorithm that calculates a Score for each Policy and categorises the policy based on the score into classes like 'Good' or 'Bad' etc. Our classes are defined and we process each paragraph of the policy starting with prepossessing the scraped data for removal of unwanted characters and summarization. We extract out the meaningful information present in each paragraph in stages and then finally classify at the last stage. Our classes are predefined and on top of that the final scoring mechanism work. 

At the implementation side of things, it becomes crucial to address,advise and aware the user of the results that the backend algorithm generate. Our chrome extension works as the first contact of user and they interact solely with the extension which intern interacts with the hosted back-end algorithm through API calls. The Chrome Extension also detects the pages which have their policies referred to different links. It scrapes data off of those pages and sends it at the required API address. It is at the top most priority to interfere with the users work at the slightest and to complete the whole process of evaluation maximizing the throughput.

In the upcoming sections we discuss the depth of the algorithms. Namely Section [$2$] discusses some related work done by others in the domain, Section [$3$] covers the implementation of the Chrome Extension, Section [$4$] covers the hosting techniques and Restful APIs that were setup, Section [$5$] covers the detailed Language Processing throught Machine Learning, Section [$6$] is about the details of the results and analysis that we did, Section [$7$] talks about the future work and Section [$8$] concludes the paper.

\section{Related Work}
At this time there are no significant solution to this \textit{people's problem} in the market. The most successful and accurate of all is a website TosDr [Terms of Service; Didn't Read] which manually generates safety ratings for all major websites, i.e. they have people who review the policies and score them and they use the scores to grade the website, but a major limitation is that there can be human errors anywhere along the way and the finitely available human resource could cater for only so many websites to be reviewed by them. Our work majorly draws inspiration from them but we tend to make the Machine Learning Algorithms that could work with any general policy on the web.

\subsection{Polisis}
An automated framework for privacy policy analysis (Polisis). They've enabled scalable, dynamic, and multi-dimensional queries on natural language privacy policies. At the core of Polisis is a privacy-centric language model, built with 130K privacy policies, and a novel hierarchy of neural-network classifiers that accounts for both high-level aspects and fine-grained details of privacy practices. Mining based application draws information from the fixed corpus of policies, incompetent with live data. Polisis’ modularity and utility is demonstrated with two applications supporting structured and free-form querying. The structured querying application is the automated assignment of privacy icons from privacy policies. The second application, PriBot, is the first freeform question-answering system for privacy policies.

\subsection{PrivacyCheck}
The two previous versions of PrivacyCheck, another chrome extension based product, incorporated the use of machine learning models to automatically answer 20 questions about the content of any given privacy policy, ten questions rooted in User Control and another ten in the European General Data Protection Regulation (GDPR). One setback was the fixed corpus of question against which the policies were evaluated, which was a certain loss of generality. The first two versions were used by about one thousand actual users over the past six years, since the first release in May 2015. In PrivacyCheck v3, they provide the capability to follow privacy policies and notify the user when policies change. Their work is the first to provide a bird’s-eye view of privacy policies to which the user has agreed.

\section{Chrome Extension}
Aimed at providing user with a easy and smooth user experience, but the key feature that we present is the automatic detection and scraping of data from the websites for analysis. To keep the notifications to the user minimum, we notify the user only when the back-end processing is done. Along with the final results, the extension shows a warning whenever user is on a page that is making them 'agree' to something, it can be a policy, which most people chose to ignore, while signing up or it can be any website that is designed to forcefully and subtly trick users into automatically agreeing to their terms by continuing on the website, some websites go to the extremes of not letting the user access the content unless they consent to their conditions. But most importantly between all these cases, what is common is that almost none of the users actually check out the policy by navigating to the required page.


\subsection{Scraping Data using the Extension}\label{SL}
The extension's scope for processing pages is the currently active page. As mentioned above the extension gets activated only if the website tries to make user agree to something and skips the gore details by providing links and ensuring the course of action of the user is to simply make a one shot agreement to their policies without detailed examination of what's within the lines. The extension scrapes of all the links on the page and looks for relevant links from within those $i.e$ the ones which take user to the \textit{Privacy Policy or Terms and Conditions} page. It then scrapes of each paragraph from there to be sent to the backend Machine Learning Model to process the policy and generate results. 

\RestyleAlgo{ruled}
\SetKwComment{Comment}{/* }{ */}
\begin{algorithm}[hbt!]
\caption{Scrape Policy Data from the page }\label{alg:two}

\SetKwProg{FCheck}{CheckPage}{:}{}
\KwData{$allLinks \gets \{\}$

        $relevantWords \gets$ \{words help identity useful links\}                        }
       {$CheckWords \gets $\{words pointing to consent\}}

\Comment*[l]{Check if the page is trying to make user agree to something} 
\eIf{$CheckPage(CheckWords)$}
    {
        \KwData{$i \gets 0;$}
                {
                    $j \gets 0; $
                    $links \gets \{\}$
                }
        $\$("a").each(function ()$  
        $ \{ $
        $    allLinks.push(this.href);$
        $ \}); $
        $ $
        
        \For{$i < allLinks.length $}
            {
                \For{$j < relevantWords.length$}
                    {
                        \eIf{$allLinks[i].includes(relevant_link_words[j])$}
                            {
                                $links.push(all_links[i]);$
                            }
                            {
                                $ j++; $
                                
                                $continue;$
                            }
                        
                    }
            }
        
        \Comment{variable 'links' contains all the links to Privacy Policy and Terms of Service Pages}
        $ $
        Call function to scrape all the data from the links:
        $scrape(links);$

    }
    {
        $exit$ \;
    }
    
 \FCheck{$CheckWords$}
    {
        \KwData{$pageText \gets \$('doc').text().split(" ")$}
                {$i \gets 0$\;}
        \For{$i < relevantWords.length$}
            {
                \eIf{$pageText.includes(relevantWords[i])$} 
                {
                    \KwRet True;
                }
                {$i++;$}
            
            }
        \KwRet False\;
    }

\end{algorithm}
The scraped data is sent to the back-end model by the extension itself using an API call (discussed in Section \ref{API}, which in-turn returns the processed policies and scores based on the algorithms discussed in Section \ref{LP}.
The extension must convey the results in a human understandable easy way which is at the very core of our work.

\begin{figure*}[t]
    \centering
     \captionsetup{justification=centering}
    \includegraphics[scale=0.5]{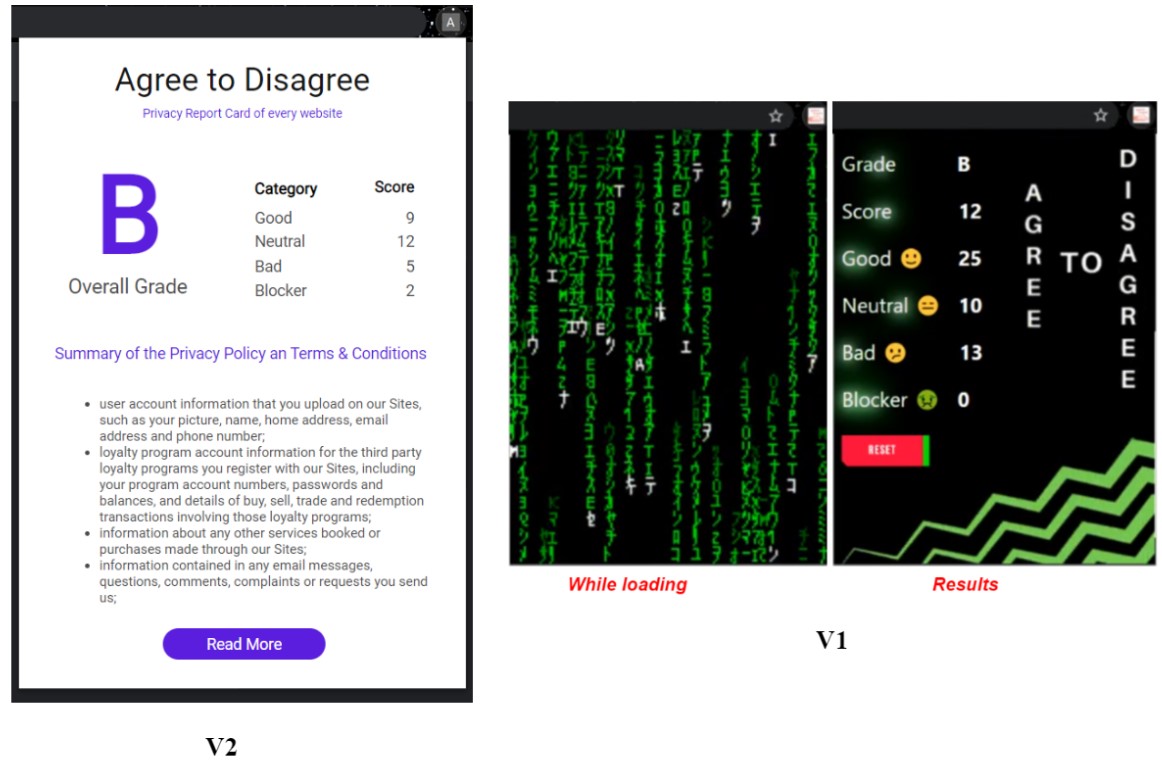}
    \caption{Version 1 (right) with just the scores for the processed policies of a website. Version 2 (left) is the latest extension which in addition also displays the summarised policies }
   \label{fig:Extension_both_versions}
\end{figure*}

As shown in the Fig \ref{fig:Extension_both_versions}, we display -
\begin{itemize}
\item \textbf{Score} : The calculated Good, Bad , Neutral and Blocker points present in the policies are shown under this column
\item \textbf{Overall Grade} : Using the scoring mechanism as mentioned in the paper [1] we score the website in a similar fashion where the good policies carry a positive relative weight and the bad ones carry a negative relative weight
\item \textbf{Summary} : Summary contains the paragraph wise summarised, easy to understand textual information retrieved from each of the paragraphs of the policy page

\end{itemize}

\subsection{Communicating with backend}\label{CWB}
After the links are obtained which contain the policies we go to each page and scrape of each paragraph of text present there. Then we send a request to a setup API which communicates with the Machine Learning Model's wrapper and in-turn does all the processing. The scraping of data from each link can be done through - 
\begin{itemize}
    \item \textbf{Injecting Payload} : In this technique we inject a payload script to the new URL which scrapes of the data and stores it as a JSON for usage.
    \item \textbf{Creating a new tab} : Here we follow a naive approach of opening a new tab with the link and scraping the data using the same script and storing the text and then immediately closing the tab automatically.
\end{itemize}

After making the call to the API and receiving the response we display the information returned by the ML model by simply switching the state of the extension from '\textit{processing}' to '\textit{processed}' and replacing the content with relevant information.

\section{Hosting Techniques and API setups}\label{API}
After scraping the data off of the policy pages we convert it to a json and send to the backend via an API call. The NLP models hosted require GPU for optimum performance but with a few added seconds out extension generated results without a GPU as well.

\RestyleAlgo{ruled}
\SetKwComment{Comment}{/* }{ */}
\begin{algorithm}[hbt!]
\caption{Send request through API}\label{alg:two}

\SetKwProg{FCheck}{CheckPage}{:}{}
\KwData{$message \gets$ JSON.stringify(data from payload); 

       $postUrl \gets$ API Url ; }  
       
$response \gets \$.post(postUrl,message); $
\Comment{response contains all the scores and summaries generated by the model}

\end{algorithm}

API can be setup using Flask on local machines.
Any virtual hosting service can be used to deploy the models that provide enough space to store heavy ( but non GPU based) models. So storage on the server is important as a requirement for the host rather than the processing speed which is an add on and use case specific.

At the receiver end we process the JSON by the extension and extract out the relevant information and display appropriate data on the extension pop up.

\begin{figure*}[t]
    \centering
     \captionsetup{justification=centering}
    \includegraphics[width=\textwidth ,height=250px]{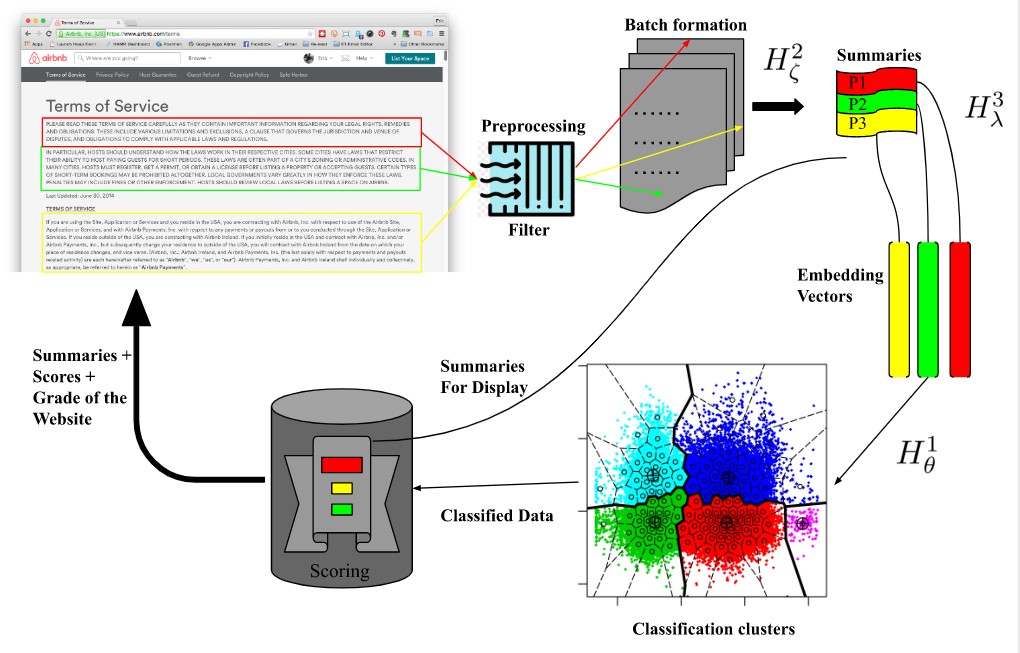}
    \caption{ Policies Scraped by the extension is pre-processed for HTML tags and other non-alphanumeric noise-like characters carefully. They are then batched up according to their chunk size.$H_\zeta ^2$ is the summarizer model. Summaries are then converted to vectors through $H_\lambda ^3$ which is the pre-trained BERT-base model, followed by $H_\theta ^1$ which is the final classifier model that is used to predict whether the policy is good or bad etc. Finally data score is generated and the summaries are passed back to the extension to display }
   \label{fig:Pipeline_overall}
\end{figure*}
\section{Language Processing}\label{LP}
In this section, we dive deep into the details of the Algorithms that work at the backend. From pre-processing the incoming data to the generation of scores there are mainly 3 models that help us in doing so-
\begin{itemize}
    \item \textbf{Summarizer}: After pre-processing the incoming policies we pass each individual chunk through a summarizer which abstractly summarises the large chunks into smaller ones.   
    \item \textbf{Tokenizer}: The BERT-based model is used to generate BERT embeddings of the summarized chunks.
    \item \textbf{Classifier}: The classifier Machine Learning Model classifies the BERT embeddings into 4 classes namely - Good, Neutral, Bad, and Blocker using which the scores are calculated. 
    
\end{itemize}

\subsection{Live Policy Evaluation}\label{OE}
Shown in Fig \ref{fig:Pipeline_overall} is the overall pipeline at the backend. The scraped policies are received at the hosted model. The policies are generally huge and due to the dynamic structures of different websites, scraping data sometimes might carry HTML/XML tags along with them. Moreover not all websites lay down their policies purely in English, according to the need and type of website we might encounter characters of other languages and some special non-alphanumeric or accented characters as well. So, our first filter tries to clean up the text, taking out all the not needed characters and converting all the text to lowercase as we will be using uncased BERT so it becomes irrelevant to have different tokens for capital and non-capital words, it can also be easily justified that the Capital words are used mostly for nouns, which usually happen to be an ineffective contributor to the overall sentiment of the legal policy. We have used \textit{BeautifulSoup's} HTML parser to get rid of the tags along with another \textit{unicodedata normalizer} that removes the accented characters making the policies ready to be summarized.

We use the \textbf{t5-base} summarizer provided by \textit{Hugging Face} pipeline respresented by $H_\zeta ^2$ in Fig \ref{fig:Pipeline_overall}. But before that, we split the message containing the policy and based on the number of words in each paragraph we summarize each small chunk. As explained in pseudocode \ref{alg:Summarize} for each chunk of variable size we accordingly summarize them into the appropriate maximum length of summaries. T5 is a large-scale transformer-based model that was trained by Google and provided by Hugging Face. T5 can perform abstractive summarization, which means that it can generate a summary of a given text by understanding the content of the text and generating a new, shorter version that conveys the most important information from the original. This is a crucial step in the pipeline as we don't want to miss out on any of the information that might decide whether the policy is good or bad. T5 summarizer would extract important information along with significantly reducing the size of the text, which makes it faster to process in the further steps of the pipeline.

\RestyleAlgo{ruled}
\SetKwComment{Comment}{/* }{ */}
\begin{algorithm}[hbt!]
\caption{Summarization}\label{alg:Summarize}

\SetKwProg{FCheck}{CheckPage}{:}{}
\KwData{$paragraphs \gets$ all the paragraphs of the policy; }
{$summarizer \gets t5-Base$;} \\  
        \For{$para$ $\in$ $paragraphs $}
            {
                \eIf{$para.length>400$}
                {$summarize(maxLength = 200)$;}{pass;}
                \eIf{$para.length>200$}
                {$summarize(maxLength = 100)$;}{pass;}
                \eIf{$para.length>100$}
                {$summarize(maxLength = 75)$;}{pass;}
                \eIf{$para.length>75$}
                {$summarize(maxLength = 50)$;}
                {pass;}
            
            }
\Comment{summaries are then stacked up in a list}

\end{algorithm}

The next step is rather simple we stack up all the summaries into a list and then we tokenize them and pass it to the \textbf{BERT} model to generate embeddings corresponding to each summary (shown by $H_\lambda ^3$ in Fig \ref{fig:Pipeline_overall}). Developed by Google, BERT is designed to pre-process text by assigning a "fixed-length" vector representation, which is what an embedding signifies it also contains the contextual information of the text in the form of a vector that we utilise for our classification task that follows.

After that, we use a Machine Learning model like KNN (represented by $H_\theta ^1$ in Fig \ref{fig:Pipeline_overall})  with $k=3$ to classify the embeddings into the 4 classes namely good, bad, blocker or neutral. Note that $H_\theta ^1$ is one Model from the family of models with trained parameters $\theta$, in general, the classifier should be the best classifier available it could also be a combination of more than one model but what should be kept in mind is that we are trying to reduce the time taken for the user to get the evaluation results so we keep things simple and fast which is our priority, but if higher accuracy is desired then heavier models can be employed. A detailed analysis of the model training is done in Section \ref{STM} meanwhile a detailed analysis of our experiments and conclusions with different ML models is done in Section \ref{R}.

The scoring algorithm can be subjective and can be modified as per the need. We assign a score as- \\
\fbox{
  \parbox{5cm}{
    $score = good - bad - 3*blocker$
    \hrule
  }
}
  \\
where $good$ carries a weight of +1, $bad$ carries a weight of -1, $blocker$ carries a weight of -3 and $neutral$ carries a weight of 0 (hence not in the equation). Then based on the final score that a website gets after evaluating and classifying each of its policies, we categorise them into different grades. We have chosen the metrics used by the website ToS;DR but the weights and scoring criteria as a whole are flexible. 

\subsection{Supervised Training Models}\label{STM}
We use 3 Models in total, 2 of them namely the T5-summarizer ($H_\zeta ^2$ in Fig \ref{fig:Pipeline_overall}) and BERT ($H_\lambda ^3$ in Fig \ref{fig:Pipeline_overall}) were made available through Hugging Face Pipelines. Both T5 and BERT are trained on a wide variety of data, including text from books, news articles, and web pages. We did not find any need to fine-tune any of them so we have deployed them directly. But the $3^{rd}$ and the most important Model of them, which is the 'classifier model'($H_\theta ^1$ in Fig \ref{fig:Pipeline_overall}) needs to be trained to predict whether a given policy can be categorized as good or bad. Once we have the embeddings vector containing the contextual information of the policies, it boils down to a very simple supervised learning task. 

\textbf{Dataset} - We have used the data from ToS;DR which aims at creating a transparent and peer-reviewed process to rate and analyse Terms of Service and Privacy Policies in order to create a rating from Grade A to Grade E. But they do this manually through human interaction.
Terms of service are reviewed by contributors and divided into small points that we can discuss, compare and ultimately assign a score with a badge. Once a service has enough badges to assess the fairness of its terms for users, a class is assigned automatically by pondering the average scores. Somewhat similar to what we do with the help of AI. We took around 4160 policies from their API which were labelled by the reviewers and assigned a score as shown in the Table \ref{tab:Dataset}.

\begin{figure}[t]
     \captionsetup{justification=centering}
    \includegraphics[scale=0.6]{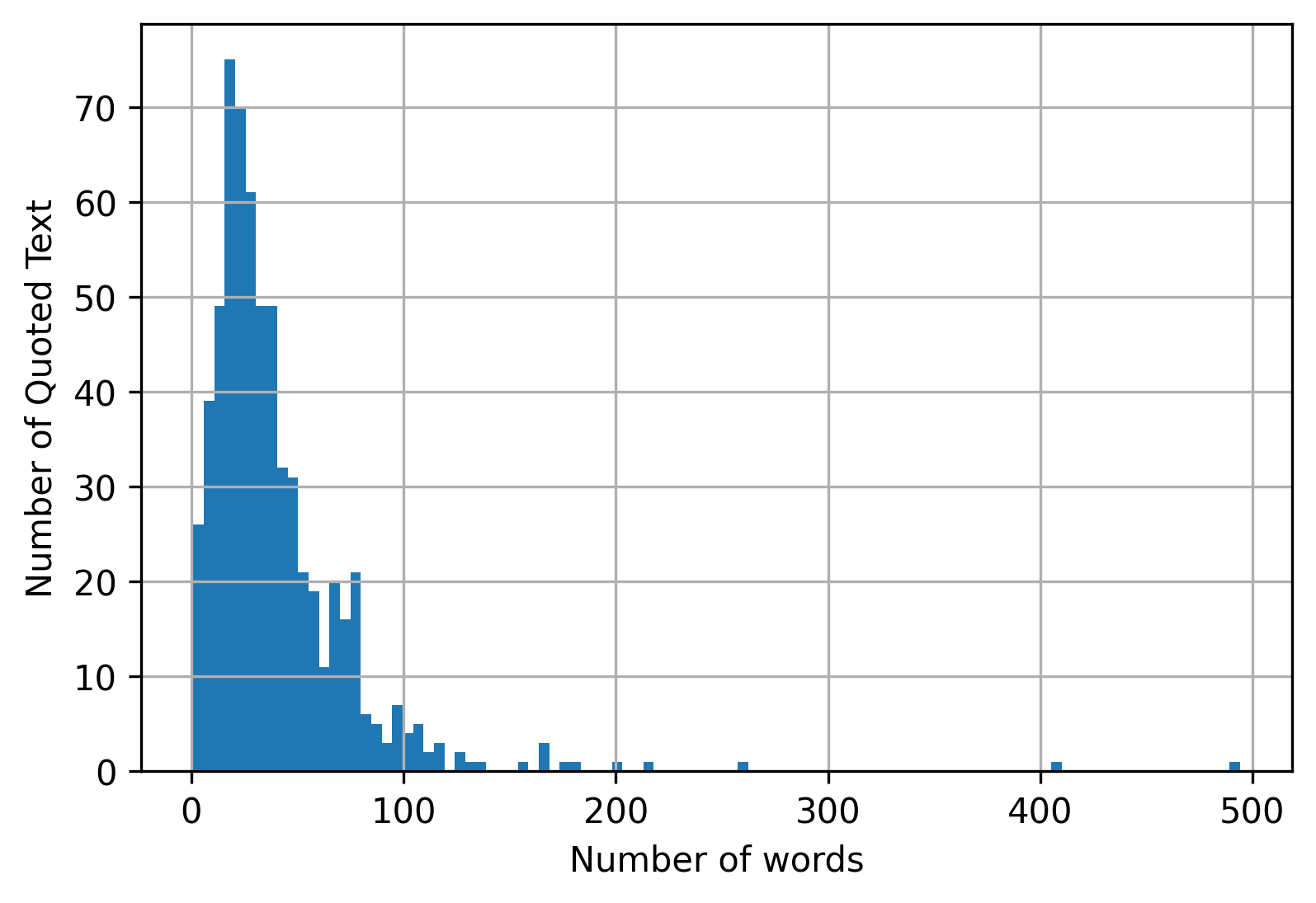}
    \caption{Plot showing the number of words in a quoted text in the dataset generated from ToS;DR's API}
   \label{fig:Dataset}
\end{figure}

Fig \ref{fig:Dataset} shows the distribution of words in each quoted text which was recieved from the ToS;DR's API. It can be concluded from the plot that even though majority of the policiy paragraphs contain around 50-60 words, there are still some policy paragraphs that contain more than 400 words as well. This is why it becomes important to summarize them as well. Summarizing them into even shorter text would make bring uniformity in terms of length of each policy since we are not losing information due to abstractive nature of summarization. It also means that while generating embeddings we would have lesser padding, since in a batch all other sentences get padded to become of the same size as that of the largest sentence in the batch, which in turn means lesser loss of information (as zeros in a vector would not be conveying any meaningful information about the policy to the ML model).

\begin{table*}[ht]

\centering
\begin{adjustbox}{width=\textwidth}
\small
\fbox{
\begin{tabular}{|l| l| l|}
\hline
  \textbf{point} & \textbf{quoteDoc} & \textbf{quoteText} \\
  \hline
  bad & Privacy Policy & Genetic Information.
</strong> Information regarding your genotype (e.g. the As, Ts, Cs,\\ & & and Gs at particular locations in your genome),your Genetic Information,  is generated \\ & & when  we analyze and process your saliva sample, or when you  otherwise  contribute or \\ & &  access your Genetic Information through our Services. \\
\hline
  good & Terms of Service & If you want to terminate your legal agreement with 23andMe, you may do so by \\ & & notifying 23andMe at any time in writing, which will entail closing your accounts for \\ & & all of the Services that you use. \\
  \hline
  neutral & Terms of Service & You should not assume that any information we may be able to provide to you, whether  \\ & & now or as genetic research advances, will be welcome or positive.\\
  \hline
 blocker & Privacy Notice & <li>search results and links, including paid listings (such as Sponsored Links). \\
 \hline
\end{tabular}
}
\end{adjustbox}
 \captionsetup{justification=centering}
\caption{Table containing data as received from ToS;DR API, note that some columns have been omitted from the actual response as they were not relevant }
\label{tab:Dataset}
\end{table*} 

As can be seen in Table \ref{tab:Dataset}, \textbf{quoteText} contains a lot of noise and the clearly the quoted document lines are not pre-processed for removing HTML tags etc, also some quoted documents are very large. We follow the same routine with the training data to make it model ready-
\begin{itemize}
    \item Quoted Document text is pre-processed to remove all non alphanumeric tokens
    \item According to the size of the doc, they are summarised in batches
    \item The summaries are stacked up and then embeddings are generated
\end{itemize}

\begin{table*}[t]
\centering
\begin{adjustbox}{width=\textwidth}
\small
\begin{tabular}{|l|l|l|l|l|l|l|}
  \hline
 & Model & Precision & Recall & F1 & Accuracy & AUC  \\ 
  \hline
  1 & KNN &  0.7155 & 0.7139 & 0.7102 &  0.7139 & 0.7736  \\ 
  2 & LSVM & 0.6511 &  0.6454 & 0.6479 & 0.6454 & 0.7367  \\ 
  3 & QDA & 0.6528 & 0.6069 & 0.5260 &  0.6069 & 0.7450 \\ 
  4 & Naive Bayes & 0.5578 & 0.5180 & 0.5335 & 0.5180 & 0.6467 \\ 
  5 & Ada-Boost & 0.5083 & 0.5168 & 0.5102 & 0.5168 & 0.6552  \\ 
  6 & Decision Tree & 0.5184 & 0.5168 & 0.5175 & 0.5168 & 0.6473 \\ 
  7 & Random Forest & 0.3637 & 0.4891 & 0.4142 & 0.4891 & 0.6155\\
    \hline
\end{tabular}
\end{adjustbox}
\caption{Model Performance} 
\label{tab:Performance}
\end{table*}

\begin{figure}[t]
     \captionsetup{justification=centering}
    \includegraphics[scale=0.6]{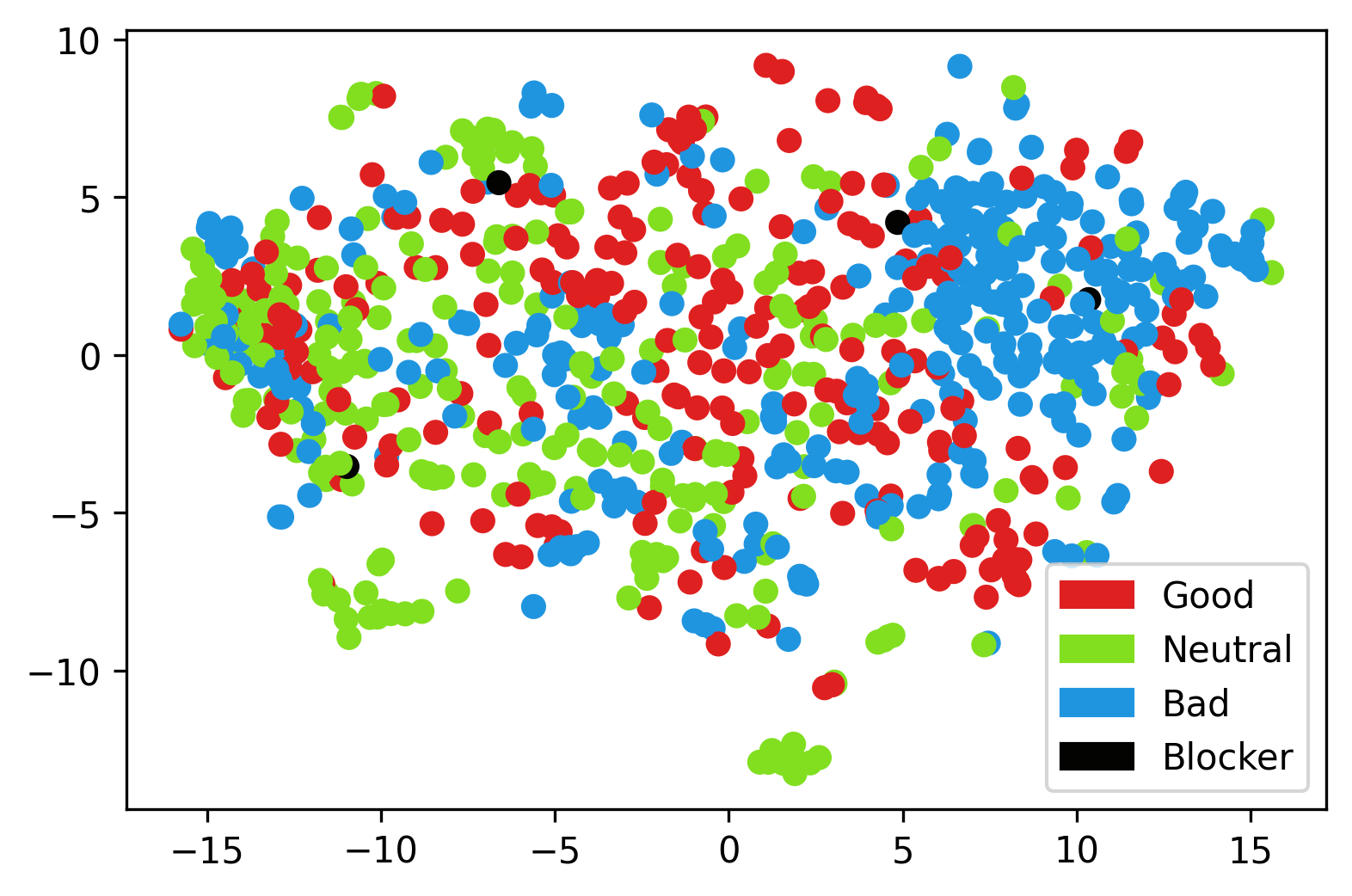}
    \caption{tSNE of embeddings representing point cluster }
   \label{fig:TSNE}
\end{figure}

\begin{figure}[t]
     \captionsetup{justification=centering}
    \includegraphics[scale=0.6]{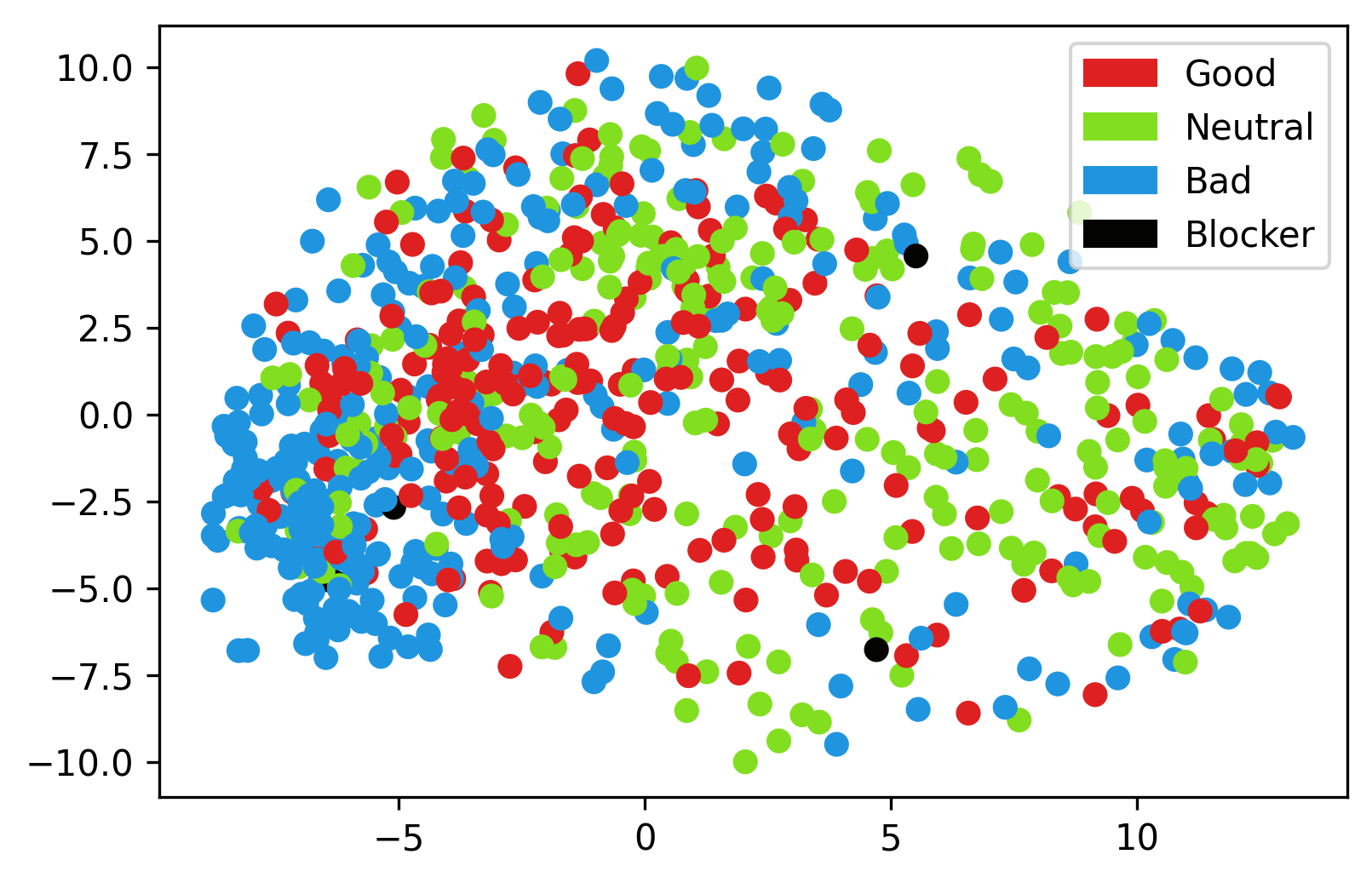}
    \caption{PCA of embeddings representing point cluster }
   \label{fig:PCA}
\end{figure}

\section{Results}\label{R}
Vectorization of policies give us the freedom to try and use the contextual meaning of the text present in the vector for various prediction and classification tasks. Since out goal is to rate a website we classified the embeddings into 4 classes.With a train test split of 80-20 \% we have tried different classifiers and the best result was shown by \textbf{KNN} with $k=3$ which gave an \textbf{accuracy of 71.39\%}. After that \textbf{Linear SVM} performed well, we achieved an \textbf{accuracy of 64.54\%}. We got \textbf{60.69 \% classification accuracy} with \textbf{Quadratic Discriminant Analysis} model provided by SkLearn. \textbf{Naive Bayes} achieved an \textbf{accuracy of 51.80 \%} very closely matched by \textbf{Ada-Boost} which could correctly classify \textbf{51.68 \%} of the test data which was also matched by the \textbf{Decision Tree} classifier. Finally the least performing model turned out to be \textbf{Random Forest} with an \textbf{accuracy of 48.91 \%}. Other accuracy metrics have been tabulated in Table \ref{tab:Performance}.

Using dimentionality reduction techniques we have reduced the each embedding in the test set to a vector in $\mathbb{R}^2$. On colouring the points in 2-D space and plotting them through a scatterplot, we can see that even after significant reduction in dimensions which is likely to cause loss of information, there are still clusters of different types formed. As shown in Fig \ref{fig:PCA} the blue points denoting the 'bad' policy points concentrate towards the left and similarly in Fig \ref{fig:TSNE} we can see the 'good' points in test set of the policy (denoted by red dots) surround the blue points. The results very well align with the model performances and make it clear that clustering is the better approach while dealing with the semantics of legal text.

\section{Future Work}
As a first attempt at real time processing of huge policies, a lot of ground has been covered but a lot more in terms of improving upon the efficiency of the overall process can be done. 
\begin{itemize}
    \item \textbf{Sampling Topics -} Some topics regarding privacy are more important than other, for ex. policies regarding your location access might be of more importance than policies regarding what browser are you using. Plan is to extract out topics and sample from some Dirichlet distribution of topics and rank them according to relevance to user and then process those parts of the policies for better analysis of the website.
    \item \textbf{Better classifiers -} Currently we take a very naive approach of classification using clustering, better methods to cluster can be employed. Exploring beyond supervised methods, unsupervised or zero shot techniques can be tried to classify on the go but keeping in mind that accuracy of prediction is of primary importance.
    \item \textbf{Speeding up scraping -} The faster the extension scrapes off the policies the more time is saved overall. Currently the code scrapes of all paragraphs, a better approach can be to selectively pick which are the relevant paragraphs to pick and send to the backend model. This would also reduce overheads at the backend pre-processing.
    \item \textbf{Speeding up API calls -} Data can be split into smaller chunks and send via multiple requests to through the API bridge and then reassembled at the backend. It is an approach similar to data packets travelling from one node to another in a network.
    \item \textbf{Worker threads at backend -} Multi-threaded backend implementation could speed up the rate at which multiple users send requests at the ML model. It can make sure that the requests don't pile up for long, and maximum efficiency rate is maintained at the backend utilizing full computational capacity of the host
    
\end{itemize}

\section{Conclusion}
Through our work we want to show that legal text does have a sentiment. The goal is to make Artificial Intelligence learn and extract out meaning from policy data, more so to help build safety evaluation systems for the future use. Not only human efforts in policy evaluation would reduce significantly we would also be able to increase public engagement in their participation in right to information and right to privacy. Privacy policies are written in legal language and often can be very misleading for individuals to understand,it easier for companies to hide potentially harmful or controversial practices in their privacy policies. As a result of which a lot of people do not actually understand how can a website potentially harm them in unknown ways. 

Growing privacy concerns across the globe calls for reliable integration of Machine Learning to ease the process of safeguarding the interests of individuals on the web. Robustness of such systems continues to be in question and more reliability is desired as we go further. Community as a whole. We present an end to end pipeline blended with a chrome extension on the front end that could help every internet user be more aware of the Terms and Conditions that he unknowingly agrees to while being on the web.





\end{document}